\documentclass[conference]{IEEEtran}
\IEEEoverridecommandlockouts
\usepackage{amsmath,amssymb,amsfonts}
\usepackage{hyperref}
\usepackage{algorithmic}
\usepackage{amsmath}
\usepackage{caption}
\usepackage{graphicx}
\usepackage{textcomp}
\usepackage{etoolbox}
\usepackage{xcolor}
\usepackage[export]{adjustbox}
\usepackage[capitalise]{cleveref}

\usepackage[nocompress]{cite}

\renewcommand{\thefigure}{\arabic{figure}}
\newcommand{\insertfig}{\renewcommand{\thefigure}{1}\vspace{15pt}\includegraphics[width=0.85\textwidth]{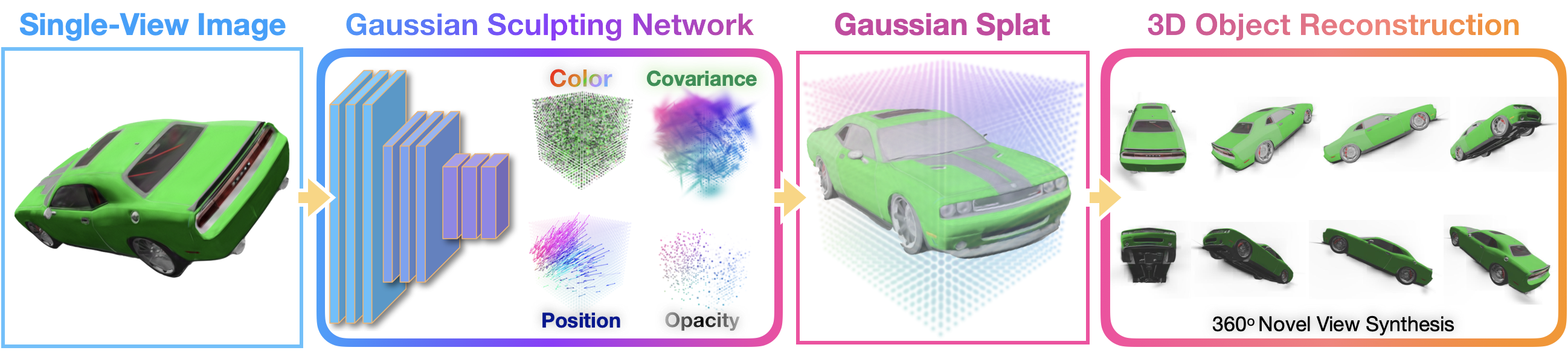}\captionof{figure}{Overview of the Gaussian Sculpting Network (GSN) for single-view 3D reconstruction. \textbf{Left:} Using a single observed image, the GSN transforms a canonical Gaussian splat with predicted colors, covariances, positions, and opacities. \textbf{Right:} Output from the GSN is combined to form a Gaussian splat representing the reconstructed 3D surface geometry and texture of the observed object.}\label{fig:teaser}\vspace{-12pt}}% define the image 

\makeatletter
\apptocmd{\@maketitle}{\centering\insertfig}{}{}% insert the figure after authors
\makeatother

\title{Single-View 3D Reconstruction via SO(2)-Equivariant Gaussian Sculpting Networks}

\author{Ruihan Xu$^{1}$, Anthony Opipari$^{1}$, Joshua Mah$^{1}$, Stanley Lewis$^{1}$\\Haoran Zhang$^{1}$, Hanzhe Guo$^{1}$, and Odest Chadwicke Jenkins$^{1}$
\thanks{$^{1}$All authors are affiliated with the Department of Robotics, University of Michigan, Ann Arbor, MI, USA, 48109.}
\thanks{$^{*}$R. Xu and A. Opipari are the corresponding authors:
        {\tt\small rhxu@umich.edu, topipari@umich.edu}}
}

\begin{document}

\maketitle

\begin{abstract}
This paper introduces SO(2)-Equivariant Gaussian Sculpting Networks (GSNs) as an approach for SO(2)-Equivariant 3D object reconstruction from single-view image observations.
 GSNs take a single observation as input to generate a Gaussian splat representation describing the observed object's geometry and texture (\cref{fig:teaser}).
By using a shared feature extractor before decoding Gaussian colors, covariances, positions, and opacities, GSNs achieve extremely high throughput ($>$150FPS).
Experiments demonstrate that GSNs can be trained efficiently using a multi-view rendering loss and are competitive, in quality, with expensive diffusion-based reconstruction algorithms.
The GSN model is validated on multiple benchmark experiments.
Moreover, we demonstrate the potential for GSNs to be used within a robotic manipulation pipeline for object-centric grasping.

\end{abstract}

\section{Introduction}
Autonomous robots tasked with operating in complex, unstructured environments must be able to perceive their environment and the objects within it.
In particular, perceiving the geometry and visual properties of objects is needed in order for robots to effectively plan and act using those objects.
At a given instant, a robot can typically only observe part of an object's surface and must rely on 3D reconstruction algorithms to perceive the remaining occluded geometric structures.
Moreover, an equivariant reconstructed representation enables to inform consistent understanding and perform reliable interaction on the objects, eg. a reliable grasp pose on the reconstructed object should not be effected by the view angle.

% Moreover, the ability to understand and interact with novel objects 

% that were previously unobserved during model training remains a critical requirement for unstructured settings.

Novel view synthesis offers a promising direction for robots to reconstruct detailed object models using only sparse viewpoints from their environment. 
For example, Neural Radiance Fields (NeRFs), is a technique that enables novel view synthesis and 3D reconstruction by using neural implicit representations to render new views of a scene based only on a small set of observed images~\cite{mildenhall2020nerf,pan2023many}.
While NeRFs have demonstrated impressive results and garnered significant interest in robotic applications, their implicit nature often hinders the explicit representation of objects' geometric structures in 3D space, posing challenges for direct integration with robotic planning and control modules~\cite{kerr2022evo,9712211,li20223d,rosinol2023nerf,lewisnerf}.

Against this backdrop, Gaussian Splatting has emerged as an explicit alternative for representing scenes and objects while also enabling novel view synthesis~\cite{kerbl3Dgaussians}.
This approach uses a collection of Gaussian-shaped primitives to approximate a continuous scene volume along with a differentiable, highly efficient rendering technique to optimize each Gaussian against the observed views.
The explicit nature of Gaussian Splatting representations allows for a simpler integration of equivariant representations, such as those based on the SO(2) rotation group, which better facilitates downstream robotics tasks.

% has the potential to be used within robotic perception by downstream planning and control algorithms.
Recent work has demonstrated the potential for Gaussian Splatting within robotic manipulation~\cite{abou2023physically,zheng2024gaussiangrasper} and navigation~\cite{keetha2024splatam,chen2024survey}.
While Gaussian Splatting has enabled novel view synthesis with an explicit representation useful for robotics, for 3D reconstruction it requires multiple diverse image observations, which are not always available.
In this project, we set out to enable high throughput 3D reconstruction using single-view images with Gaussian Splatting, aiming to enable robots to understand the geometry and texture of unseen parts of objects consistently across different view angle. 

The present paper makes the following contributions:
\begin{enumerate}
    \item Propose and develop SO(2)-Equivariant Gaussian Sculpting Networks (GSNs), to generate Gaussian splats from single-view image observations in real time. 
    \item Perform quantitative and qualitative evaluations of GSNs to understand relative tradeoffs in comparison to existing 3D reconstruction approaches.
    \item We demonstrate the potential for GSN to be used within a robotic manipulation pipeline for grasping.
\end{enumerate}

\renewcommand{\thefigure}{2}
\begin{figure*}[ht!]
    \centering
    \includegraphics[width=\linewidth]{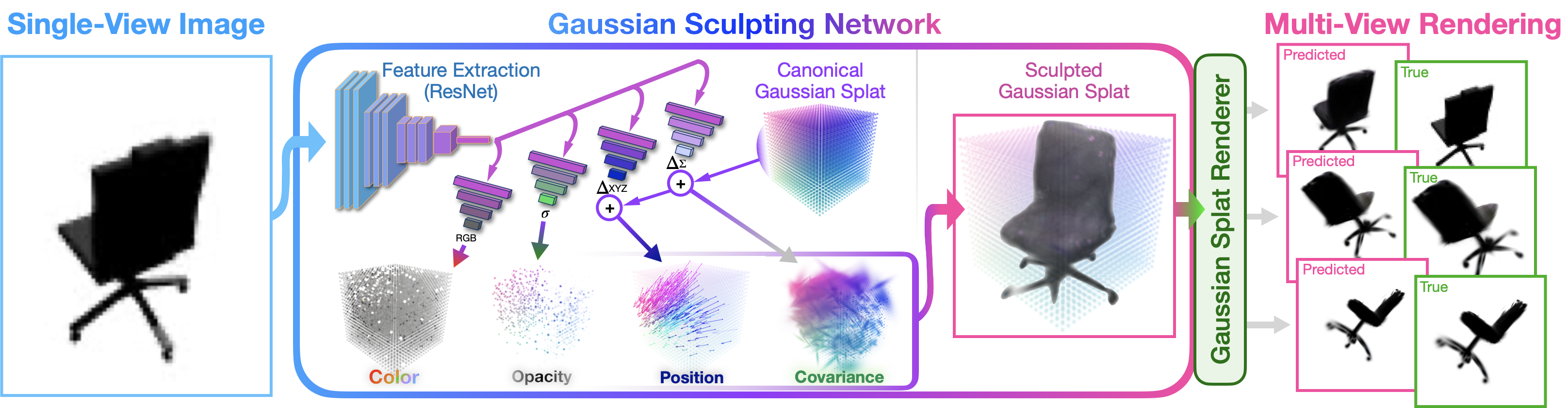}
    \caption{SO(2)-Equivariant Gaussian Sculpting Network Architecture. Our Encoder-Decoder style network takes an input image and encodes it into a latent vector. Subsequently,
a decoder with parallel MLPs decodes the latent vector into Gaussian parameters, sculpting a canonical Gaussian Splat into a 3D object presented in the input
image. Finally, we perform multi-view rendering to obtain various novel views for loss calculation.}
    \label{fig: architecture}
\end{figure*}

\section{Related Work}

\renewcommand{\thefigure}{2}

\subsection{Multi-View 3D Reconstruction}

Multi-view 3D reconstruction is a foundational topic within computer vision and robotic perception that seeks to infer the three-dimensional structure of a scene using numerous images taken from different vantage points throughout the scene.
Classical approaches, such as Structure from Motion (SfM) and Multi-View Stereo enable the derivation of 3D points or volumetric representations by triangulating matched features across multiple images~\cite{Hartley2000,1640800}.
End-to-end deep learning-based approaches have been considered including MVSNet~\cite{yao2018mvsnet} and DeepMVS~\cite{huang2018deepmvs} for depth map estimation as well as DeepSDF~\cite{park2019deepsdf} as a continuous shape representation.
More recently, neural radiance fields (NeRFs) are proposed as an approach to implicitly reconstruct volumetric scenes from sparse image observations~\cite{mildenhall2020nerf,reizenstein2021common,pan2023many}.
Following the surge of interest in NeRFs, Gaussian splatting is developed as an explicit approach to novel view synthesis from multi-view images~\cite{kerbl3Dgaussians}.
Notably, Gaussian splatting represents a scene as a mixture of Gaussians distributed throughout the scene and optimized using gradient descent.
In contrast to these approaches, the present paper sets out to reconstruct 3D scenes using only a single observed image.

\subsection{Single-View 3D Reconstruction}
In single-view 3d reconstruction, algorithms have access to only a single image of the object whose shape they set out to estimate.
Early approaches used shading cues to estimate geometric surface contours~\cite{horn1970shape,DUROU200822}.
Following the rise of deep learning, convolutional neural networks have been proposed for voxel-grid reconstruction from single images~\cite{yan2016perspective}.
Image-conditioned reconstruction into point cloud representations has been proposed using convolution~\cite{yu2018pu} and diffusion-based architectures~\cite{zhou20213d,nichol2022pointe}.
Following the introduction of Gaussian splatting, 3D reconstruction using splats as a 3D representation has gained interest~\cite{szymanowicz2023splatter,watson2022novel,feng2024fdgaussian}.
Watson et al. introduce 3DiM to generate novel views with a single image using image-based diffusion, then generate 3D splats using those generated images~\cite{watson2022novel}.
Similarly, Feng et al. employ image-level diffusion with additional geometric constraints to improve the reconstruction speed and quality.
Szymanowicz et al. propose a convolutional architecture to directly regress an output Gaussian Splat representation offering substantial speedups~\cite{szymanowicz2023splatter}.
We set out to further improve the efficiency of single-view reconstruction based on Gaussian splatting.

\vspace{-4pt}
\section{Method}

\subsection{Gaussian Primitive: 3D Gaussian Splatting}\label{subsection:gaussian_primitives}

A Gaussian Primitive \(\mathcal{G}\) is parameterized by 4 variables: mean \( \mu \in \mathbb{R}^3\), covariance  \( \mathbf{\Sigma} \in \mathbb{R}^{3 \times 3} \), RGB color \( c \in \mathbb{R}^3 \), and opacity \( \alpha \in \mathbb{R}\).
To simplify optimization, we follow~\cite{kerbl3Dgaussians} and decompose \(\mathbf{\Sigma}\) into scale \(s \in \mathbb{R}^3 \) and rotation \(r \in \mathbb{R}^4 \) represented using rotation quaternions.
The resulting Gaussian will have the form of \( \mathcal{G} = (\mu, s, r, c, \alpha)\). A scene or object can thus be represented by a collection of Gaussians that form a Gaussian Splat \[\mathcal{S} := \{\mathcal{G}_i, i = 1,..., N \} =  \{(\mu_i, s_i, r_i, c_i, \alpha_i), i = 1,..., N \}\]

The 3D Gaussian Splatting\cite{kerbl3Dgaussians} renderer \(\mathcal{R}\) takes a 3D Gaussian Splat \(\mathcal{S}\) and maps it into a 2D image \(\mathcal{I}\) using camera pose \(p \in \mathbb{R}^3\) and rotation \(\mathbf{R} \in \mathbb{R}^{3 \times 3}\).
Here, we want to find the function \(\mathcal{F}\) that inverts the process \(\mathcal{R}\) such that \(\mathcal{F}(\mathcal{I}) = \mathcal{R}^{-1}(\mathcal{I}) = \mathcal{S}\).
The reconstruction from this single image should preserve the correct 3D representation of the object, allowing it to render novel views of the object from viewpoints different to the input image. 

Instead of letting the neural network directly predict the 5 parameters for each Gaussian, our method predicts a deviation of the parameters from a canonical cube, similar to \cite{park2021nerfiesdeformableneuralradiance}.
The canonical cube refers to a collection of Gaussians, \(\Bar{\mathcal{G}}_i\), that are evenly spaced apart on a unit cube with fixed initial covariances \(\Bar{\mathbf{\Sigma}} \). Experimentally, this value of cavariance is initialized as an identity matrix mutiplied by a small constanct, ie. \(\Bar{\mathbf{\Sigma}} = 0.0075 \times I \). And, color and opacity values that are randomly sampled without any constraints.
The resultant initial Gaussian Splat would thus have predefined \((\Bar{\mu_i}, \Bar{s_i}) \; \forall i = 1,\hdots, N\) and would resemble a point cloud since the covariance scale is isotropic in \(\mathbb{R}^3\). 
To generate the predicted Gaussian Splat \(\mathcal{S}_{pred}\), our network \(\mathbf{f}\) predicts an deviation term \(\Delta\mu_i, \Delta s_i\) and values \(r_i, c_i, \alpha_i\) for a fixed number of Gaussians \(N\) such that
\[ \mathbf{f}(\mathcal{I}) = \{(\Delta\mu_i, \Delta s_i, r_i, c_i, \alpha_i), i = 1,\hdots, N\}\] 
The final predicted Gaussian Splat has the form 
\[\mathcal{S}_{pred} = \{(\Bar{\mu_i} + \Delta\mu_i, \Bar{s_i} + \Delta s_i, r_i, c_i, \alpha_i), i = 1,\hdots, N\} \]

\renewcommand{\thefigure}{3}
\begin{figure}
    \centering
    \includegraphics[width=\linewidth]{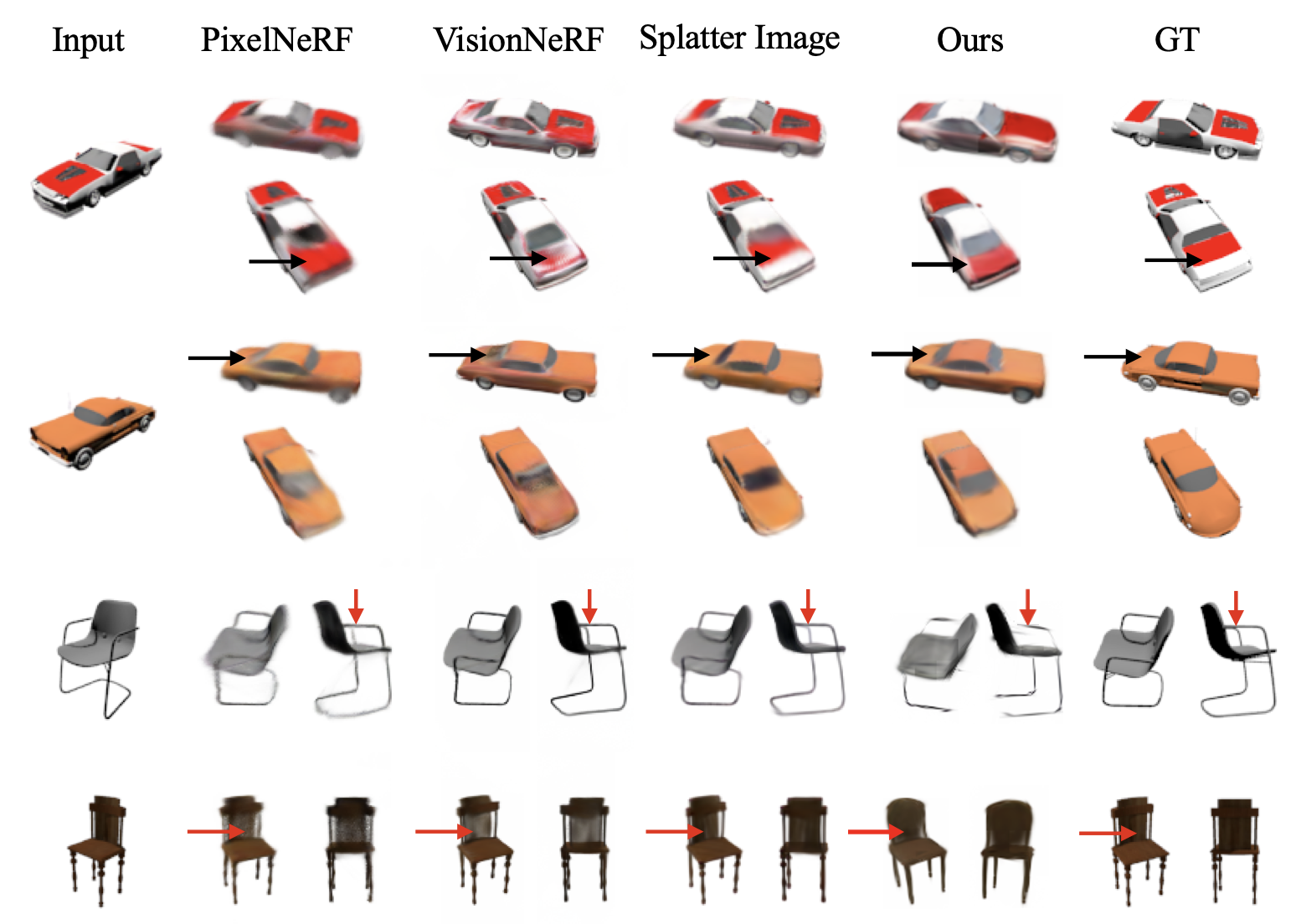}
    \caption{Qualitative single-view reconstruction comparison. We reused the visualization results from Splatter-Image and added our results for comparision.}
    \label{fig: qualitative results}
    \vspace{-1mm}
\end{figure}

\subsection{Network Architecture}

As illustrated in~\cref{fig: architecture}, the network feature extractor takes the image as input and encodes it into a latent vector followed by a decoding step into Gaussian parameters. 

\subsubsection{Encoder}
We adopted ResNet \cite{resnet} as the backbone of the encoder with some modifications.
Instead of directly using the 2048 size latent vector output from the ResNet, we take the output from the second last layer that has a size of \(2048 \times 7 \times 7\) latent tensor.
This tensor is then fed into two 2D convolutions followed by ReLU activation and finally gets a vector of size \(1 \times 2352\) after flattening the tensor.
The design was chosen since, experimentally, taking the output from the ResNet with a smaller receptive field allows the network to capture more fine-grained details to reconstruct the Gaussian Splat. We discuss this choice in more detail in the ablation section \ref{subsubsection:ablation}

\subsubsection{Decoder} \label{subsubsection:decoder}
Parallel MLPs are used to decode the latent vector into Gaussian parameters \(\Delta S = \{(\Delta\mu_i, \Delta s_i, r_i, c_i, \alpha_i), i = 1,\hdots, N\}\).
For each of the 5 parameters, we have an independent MLP layer to output \(N \times size\ of\ parameter\) number of values, where N is predefined as the number of Gaussian.
For example, to describe \(\Delta\mu\) for one Gaussian, we need three values, \(\Delta\mu_x, \Delta\mu_y, \Delta\mu_z\), so the output of the MLP for \(\Delta\mu\) will have \(N \times 3\) output size.
For each MLP, there consists of two fully connected linear layers and the first linear layer is followed by a ReLU activation.
For predicted scale \(\Bar{s}_i+\Delta s_i\), we apply ReLU activation followed by addition with a small scalar constant \(\epsilon \in \mathbb{R} : \epsilon > 0\) to ensure that the covariance is positive definite.
For predicted rotation \(r_i\), we normalize \(r_i\) without making the real quaternion term positive, whereas for color \(c_i\) and opacity \(\alpha_i\), we apply a channel-wise sigmoid function.

\begin{table}
\captionsetup{font=scriptsize}
\resizebox{\columnwidth}{!}{
    \centering
    \begin{tabular}{c|c c c |c c c}
    Method & \multicolumn{3}{c|}{Chairs} & \multicolumn{3}{c}{Cars}\\ 
      & PSNR $\uparrow$ & SSIM $\uparrow$ & Lpips $\downarrow$ & PSNR$\uparrow$ & SSIM$\uparrow$ & Lpips $\downarrow$\\
    \hline
    SRN & 22.89 & 0.89 & 0.104 & 22.25 & 0.88 & 0.129 \\
    CodeNeRF & 23.66 & 0.90 & 0.166 & 23.80 & 0.91 & 0.128 \\
    PixelNeRF & 23.72 & 0.90 & 0.128 & 23.17 & 0.89 & 0.146 \\
    \hline
    Splatter & \textbf{24.43} & \textbf{0.93} & \textbf{0.067} & 24.00 & 0.92 & \textbf{0.078}\\
    GSN (ours) & 24.35 & 0.90 & 0.117 & \textbf{24.12} & \textbf{0.92} & 0.100\\
    \hline
    \end{tabular} 
    }
\caption{\textbf{ShapeNet-SRN: Single-view 3D reconstruction.} Our model is trained on a machine with RTX A6000. The statistics for other methods are adopted from experiments in prior works. The result shows that our model performs better than some of the prior methods, while remain comparable for similar works that reconstruct a Gaussian Splat.}
\label{tab: quantitative results}
\vspace{-4mm}
\end{table}

\begin{table}
    \centering
    \captionsetup{font=scriptsize}
    \begin{tabular}{c| c }
    Method & FPS$\uparrow$  \\
    \hline
    Splatter  & 50  \\
    GSN (ours) & \textbf{164}  \\
    \hline
    \end{tabular}
    \caption{Performance during inference. Our method is significantly faster than the recent baseline for generating Gaussian Splat from a single input image.}
    \label{tab: hardsare results}
    \vspace{-4mm}
\end{table}

\subsection{Loss Formulation}
To train the neural network \(\mathcal{F}\), we aim to minimize the difference between a set of ground truth images with different camera poses and the rendered images from a predicted Gaussian Splat.
Specifically, given an input image of an object, the network predicts a 3D Gaussian Splat conditioned on this input image.
We then render the Splat from different views and compare it with available ground truth images taken from those viewpoints.
\begin{equation}
\nonumber
\begin{split}
\Tilde{\mathcal{I}_i} := \mathcal{R}(\Tilde{\mathcal{S}}, p_i, \mathbf{R}_i)\\
\mathcal{I}_i = \mathcal{R}(\mathcal{F}(\mathcal{I}_{input}), p_i,  \mathbf{R}_i)\\
\mathcal{L}_{image} = \frac{1}{k}\sum_{i=1}^k f_{image}(\Tilde{\mathcal{I}_i}, \mathcal{I}_i)\\
\end{split}
\end{equation}
where \(\Tilde{\mathcal{I}_i}\) denotes the ground truth image i, \(\Tilde{S}\) denotes some hypothetical ground truth Gaussian Splat that would generate \(\Tilde{\mathcal{I}}\) and \(k\) represents the number of sampled ground truth images with different camera poses for each object.

We adopt 3 different losses: L2, Lpips, and DSSIM as inspired by \cite{szymanowicz2023splatter}, \cite{kerbl3Dgaussians}, and \cite{lpips} which shows that L2 and DSSIM contribute to shaping the overall structure and color and Lpips help learn more fine-grained details.

To achieve SO(2) invariance regarding the view angle, the input image is rotated along the camera's principal axis by a random angle \(\mathbf{T}\) and then processed by the same network to produce a rotated splat. We minimize an Extended Chamfer Distance (ECD) between the rotated splat and the unrotated splat after applying the same rotation around the camera's principal axis.
\begin{equation}
\nonumber
\mathcal{L}_{rot} = \mathcal{L}_{Chamfer}(\mathbf{T}(\mathcal{F}(\mathcal{I}_{input})), \mathcal{F}(\mathbf{T}(\mathcal{I}_{input})))
\end{equation}

The Extended Chamfer Distance, similar to the well-known Chamfer Distance for point cloud, applies additional element-wise L2 loss on color, scale, rotation, and opacity for the nearest neighbors 

% The total loss is a combination of all 3 losses. Our specific loss formulation comes from a combination of the Splatter Image loss \cite{szymanowicz2023splatter} and 3D Gaussian Splatting loss \cite{kerbl3Dgaussians}.

% % \begin{equation}
% % \begin{split}
% % \nonumber
% % f_{loss} &:= (1 - \lambda_{lpips} - \lambda_{DSSIM}) \times \mathcal{L}_{L2} \\ 
% % &+ \lambda_{lpips} \times \mathcal{L}_{lpips}  + \lambda_{DSSIM} \times \mathcal{L}_{DSSIM} 
% % \end{split}
% % \end{equation}
% Qualitatively, we note that L2 and DSSIM\cite{baker2023dssim} seemed to contribute to shaping the overall structure and color of the object.
% Using only these losses, the final output would appear fuzzy, a phenomenon noted in~\cite{lpips}, hence lpips were added to help learn more fine-grained details.

% In practice, we randomly sample 3 other images on top of the original input image to compute the loss for each data point.

% The original input image would be used to assess how well the model can encode and reconstruct observable object details while the final 3 images would capture how well the model can generalize for unseen portions of the object.

\renewcommand{\thefigure}{4}
\begin{figure}[ht!]
    \centering
    \includegraphics[width=\linewidth]{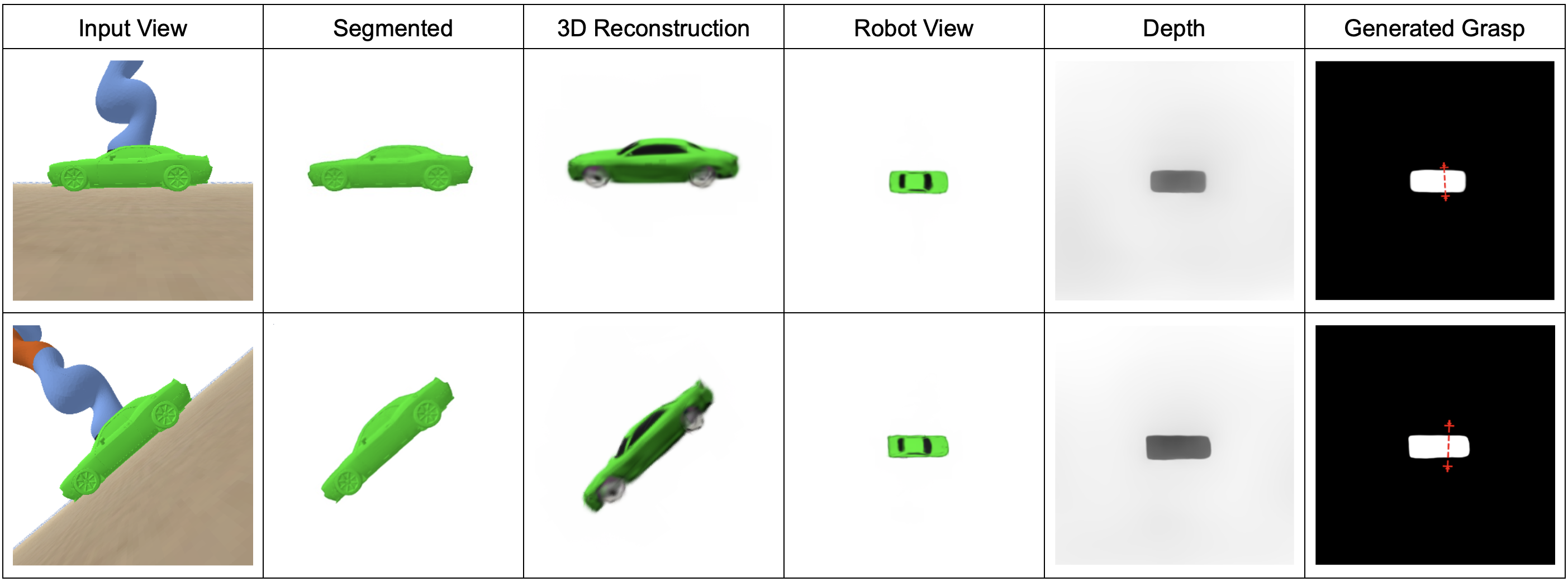}
    \caption{Generated grasp using Dex-Net 2.0 from novel views rendered from 3D Gaussian Splat generated by our model with canonical view and rotated view as input. Dex-Net can generate grasps for for cars with grasp quality q-vralue=0.733 with canonical view and q-value=0.614 for rotated view.}
    \label{fig: grasp generation}
\end{figure}

\vspace{-4mm}
\section{Experiment}

To evaluate our proposed framework, we trained our model against the chairs and cars dataset provided by~\cite{szymanowicz2023splatter}, and demonstrate possible usage for robot grasping task. 

\subsection{Single-View 3D reconstruction on Chairs and Cars}

In this task, the model predicts a 3D representation (Gaussian Splat in this case) of an object given a single input view of that object.
Novel target views are then rendered from this representation and the performance is measured by comparing these rendered images with the ground truth images.
We follow the practice of prior works, such as \cite{pixelnerf} \cite{mildenhall2020nerf} \cite{srn} \cite{vit_single_3d}, and measure the performance in terms of the Peak Signal-to-Noise Ratio (PSNR), Structural Similarity (SSIM) and perceptual quality (LPIPS).
For PSNR and SSIM, higher values indicate better quality, whereas for LPIPS, lower values are preferable. For datasets, we adopt the ShapeNet-SRN dataset for cars and chairs from \cite{srn} which aligns with the recent single-view Gaussian Splat generation model \cite{szymanowicz2023splatter}. 

\textbf{Results.} We trained our model on the chair and car datasets for four days, using a batch size of 8, until convergence. Compared to the Splatter Image model \cite{szymanowicz2023splatter} that requires seven days to converge, our model converges faster and is less expensive to train.
Table \ref{tab: quantitative results} and \ref{tab: hardsare results} and Figure \ref{fig: qualitative results} show how our model performs quantitatively and qualitatively to other baselines.
As demonstrated, our method performs roughly 3 times faster than the next fastest method (splatter-image) during inference while having comparable quantitative image reconstruction quality to other recent baselines. To investigate the reason for the faster inference, we compared the number of parameters in both GSN and Splatter-image. Splatter-image has approximately 56 million parameters, while each parallel MLP path of GSN contains about 55 million parameters. Although the number of parameters is relatively similar, GSN primarily involves linear operations (fully connected layers), whereas Splatter-image requires convolution operations. Therefore, we suspect that this architectural difference contributes to the observed speed difference. Further studies on timing the operation speed is needed to justify this hypothesis. 
Qualitatively, we observe that while our model accurately captures the overall geometry, it still struggles with extremely fine-grained details, especially in the chair dataset, which includes many thin leg and handle structures. This also explains the larger disparity in LPIPS scores for the chair dataset, as LPIPS focuses more on details, whereas PSNR and SSIM consider more of the geometric similarity.

\vspace{-6pt}
\subsection{SO(2)-Equivariant Grasp Simulation}
We also evaluate the effectiveness of our model for practical robot manipulation tasks to gauge its suitability for real-world applications. In our experiment, a Kuka arm is used along with an external camera placed on the opposite side of the robot in the Bullet simulation \cite{bulletsim}. In addition, the input view is randomly rotated along the camera's principal axis, representing imperfect view angle to test the the ability of the model to represent the SO(2) invariance with same input image feature. This setup reflects practical robot manipulation scenarios where the camera cannot observe the geometry of the object that the robot can act on. Without ground truth observation from the robot's viewpoint, GSN becomes useful in predicting the geometry and texture of the object's occluded parts, enabling the robot to generate a reliable grasp.

Specifically, as shown in Figure \ref{fig: grasp generation}, the model receives one input image and reconstructs the full object in Gaussian Splat representation. This Gaussian Splat is then rendered into a view from the robot's perspective, such as a bird's-eye view. Based on this robot view, we perform depth estimation using MiDaS \cite{Midas} to generate a depth map and Segment Anything (SAM) \cite{kirillov2023segment} to produce a binary mask. Finally, the rendered robot view, depth map, and binary mask are utilized for robot grasp prediction by Dex-Net 2.0 \cite{mahler2017dexnet}. To generate a robot motion plan to achieve the grasp pose, we use the inverse kinematics solver in PyBullet to control the Kuka arm. The simulation result is shown in Figure \ref{fig: grasp_video}, capturing snapshots at five different stages during grasping.

\renewcommand{\thefigure}{5}
\begin{figure}
    \centering
    \includegraphics[width=\linewidth]{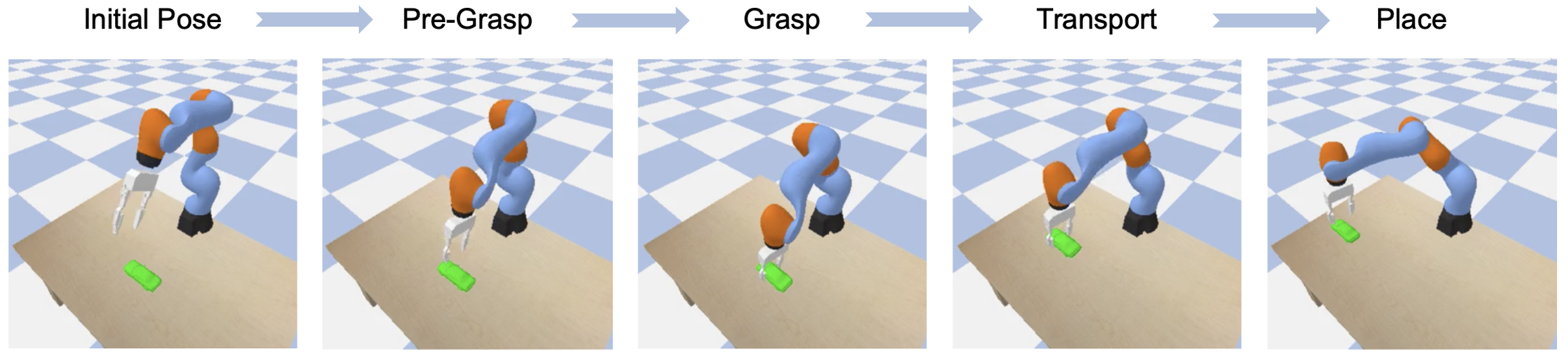}
    \caption{Simulationof Robot Grasping using Kuka manipulator in Pybullet. Check out the video demo here: \url{https://youtu.be/USOLfgzOTf4}}
    \label{fig: grasp_video}
    % \vspace{-4mm}
\end{figure}

\begin{table}
    \centering
    \captionsetup{font=scriptsize}
    \scriptsize
    \begin{tabular}{c|c|c|c}
         & ground truth & w/o rotation & with rotation\\
         \hline
       Q-value \(\uparrow\) & 0.958 & 0.957 & 0.957  \\
       \hline
    \end{tabular}
    \caption{Grasp quality of predicted grasp acted on the ground truth top-down view. The higher the better}
    \label{tab:grasp quality}
\end{table}

We observed that our model is able to represent the same object almost concisely across different view angle. In Figure \ref{fig: grasp generation}, the rotated view produce the same Splat representation and results in consistent depth estimation and grasp generation. In addition, we found that our model capture the general geometry and texture even when the input view of the object lacks rich information. Specifically, due to the compatibility issues with the object mesh, the Shape-Net car does not display the color of the windows and wheels. However, our model accurately recovers the correct color of these parts. 

In addition, we compare the grasp quality in scenarios with and without a top-down camera that directly observes the ground truth top view of the object. Following the above steps, we generate and store grasps for two different cases: side-view of the object with and without rotation. Next, we create a top-down simulation camera to capture the ground truth bird's-eye view of the object and predict the grasp and grasp quality based on this ground truth information. We then evaluate the grasp quality of the two generated grasps using the ground truth bird's-eye view and compare them with the ground truth grasp quality, as shown in Table \ref{tab:grasp quality}. Our results indicate that the Q-value remains comparably consistent across the three different settings, demonstrating that our model can generate reliable grasps even without the ground truth view of the object part the robot aims to grasp.

\section{Ablation} \label{subsubsection:ablation}

\subsection{Architecture Encoder Choice} 

In order to design the network to achieve the best output quality, we consider three key areas: the size of the ResNet architecture used, the selection of the latent vector, and the selection of the origin of the unit cube used to generate the initial set of Gaussian. We compare between ResNet50 and ResNet152 for the ResNet Architecture. For the latent vector, we consider either taking the final output of the ResNet (size 2048) or the second last layer (size 100352) and compressing it with 2 convolution layers (size 2352), ensuring the receptive fields of the latent vector in either case to be roughly similar. For the cube origin we consider one where the camera is centered on the origin and the cube 1 meter ahead from the camera, and the other where the unit cube is on the origin and the camera is 1 meter away from the origin. In both cases, the Gaussian splat has been rotated so that the camera remains upright in both frames of reference. This implies that the deformation of the Gaussians on the unit cube directly encodes the 3D shifts
required to capture the shape of the object with respect to the
camera viewpoint.

\textbf{Results}: For each of the scenarios listed in Table \ref{tab:comparative study}, we trained with a smaller data set containing 30 car objects with 150 views each. We use LPIPs as the quantitative measure. Table III shows the results of the comparative study. The greatest impact comes from centering the unit cube on the origin of the frame. The low impact of the latent vector and architecture might however be due to the small dataset and further studies would be conducted given more time

\begin{table}
    \centering
    \captionsetup{font=scriptsize}
    \scriptsize
    \begin{tabular}{c|cc|cc}
       Method  & \multicolumn{2}{c|}{ResNet50}  & \multicolumn{2}{c}{ResNet152} \\
        & Train & Val & Train & Val \\
       \hline
        Zero-Centered (2048) & \(4.37e^{-3}\) & \(5.44e^{-3}\) & \(4.30e^{-3}\) & \(5.32e^{-3}\) \\
        Off-Centered (2048) & \(5.07e^{-3}\) & \(5.64e^{-3}\) & \(5.59e^{-3}\) & \(6.07e^{-3}\)\\
        \hline\hline
        Zero-Centered (2352)* & \(3.8e^{-3}\) & \(5.35e^{-3}\) & \(4.02e^{-3}\) & \(4.68e^{-3}\) \\
        Off-Centered (2352)* & \(4.26e^{-3}\) & \(5.58e^{-3}\) & \(4.68.37e^{-3}\) & \(5.64e^{-3}\) \\
        \hline
    \end{tabular}
    \caption{Lpips results for comparative studies of architecture choice}
    \label{tab:comparative study}
\end{table}

\subsection{Effect of Extended Chamfer Distance} 

To study the effectiveness of Extended Chamfer Distance (ECD) loss, we train our model with and without this newly constructed loss on the cars dataset for 20 epochs. In addition to PSNR, SSIM, and Lpips, we also evaluate the ECD of the output splat when the input images are rotated. Shown in Table \ref{tab: Chamfer effect}, with ECD, the PSNR, SSIM, and Lpips remain comparably the same with a small performance drop, but there is a significant improvement in preserving the equivariance property when the input images are rotated, as ECD shows a huge decrease on the model trained for equivariant purposes. 

This study demonstrates that the ECD is effective in capturing the equivariant properties of the data. However, further investigation is necessary to understand the impact of ECD on reconstruction quality. Developing an improved version of ECD could be a future direction to achieve equivariance without compromising reconstruction quality.

\begin{table}
\captionsetup{font=scriptsize}
\resizebox{\columnwidth}{!}{
    \centering
    \scriptsize
    \begin{tabular}{c|c c c c }
    Method & \multicolumn{4}{c}{Cars}\\ 
     & PSNR $\uparrow$ & SSIM $\uparrow$ & Lpips $\downarrow$ & ECD $\downarrow$\\

    \hline
    GSN* w/o ECD & 23.27 & 0.91 & 0.120 & 24.22 \\
    GSN* w/ ECD & 22.61 & 0.90 & 0.125 & \textbf{3.51}  \\
    \hline
    \end{tabular} 
    }
\caption{Study of how the Extended Chamfer Distance helps the model to learn shape, texture, and equivariance. *Models are trained with only 20 epochs on single object dataset for evaluation.}
\label{tab: Chamfer effect}
\vspace{-4mm}
\end{table}
\vspace{-4pt}
\section{Conclusion}

This paper makes three central contributions: (1) a SO(2)-Equivariant Gaussian Sculpting Network for real-time single-image 3D reconstruction with the new Extended Chamfer Distance, (2) experiments validating the sculpting network design, and (3) demonstrations to highlight how GSN is applicable to a robotic grasping pipeline and remain consitent under different SO(2) transformation along the camera's principal axis. 
Experiments demonstrated the high-throughput of GSNs and suggest their potential for application to robotic perception tasks, especially in a cluttered scene where multiple objects reconstructions are required. 

GSN has several limitations that can be addressed by future work. First, GSN is currently trained only on ShapeNet synthetic datasets with limited variety, resulting in poor performance when generalizing to other synthetic or real-world data. However, since GSN only requires dense RGB images of objects, various datasets such as CO3D \cite{reizenstein2021commonobjects3dlargescale} and YCB \cite{Calli_2015ycb} can be used directly to train GSN. Future work can focus on leveraging multiple datasets to enhance GSN's generalizability. Second, GSN is an object-centric model and does not account for noise in input images, making it unsuitable for scene-level tasks. Future directions may include scene-level single-image reconstruction or multiple object reconstruction in cluttered scenes with heavy occlusion. Third, although the current design of the Extended Chamfer Distance (ECD) helps GSN learn equivariance, it involves a trade-off with reconstruction quality. Therefore, a better version of ECD could be developed to achieve both high reconstruction quality and good equivariance preservation.

Overall, our results highlight the potential of using Gaussian Splatting as the primitives for object-centric perception with equivariance awareness, facilitating downstream robotic tasks such as object manipulation.
\newpage
\clearpage
{
    \small
    \bibliographystyle{ieeetr}
    \bibliography{main}
}

\end{document}